\documentclass{article}
\usepackage{ijcai17}

\usepackage{times}
\usepackage{proof}  
\usepackage{bussproofs}
\usepackage{verbatim}

\title{On Automating the Doctrine of Double Effect}
\date{\today}
\author{Naveen Sundar Govindarajulu \normalfont{and}
\textbf{Selmer Bringsjord}\\ 
Rensselaer Polytechnic Institute, Troy, NY  \\
\{naveensundarg,selmer.bringsjord\}@gmail.com}

\usepackage{url}

\usepackage{pslatex}
\usepackage{hyperref}

\usepackage{booktabs}
\usepackage{color}
\usepackage{amsfonts}
\usepackage{amsmath}
\usepackage{proof}
\usepackage[amsmath,thmmarks]{ntheorem}
\usepackage{relsize}
\usepackage{mathtools}    
\DeclarePairedDelimiterX\setc[2]{\{}{\}}{\,#1 \;\delimsize\vert\; #2\,}
\newcommand{\set}[1]{\left\lbrace #1\right\rbrace}

\newcommand{\IGNORE}[1]{}
\newcommand{\newtext}[1]{#1}

\newcommand{\lsort}[1]{%
  \ensuremath{\mbox{\textsf{#1}}}}
\newcommand{\defsort}[2]{%
  \newcommand{#1}{\lsort{#2}}}
\defsort{\Action}{Action}
\defsort{\Time}{Time}
\defsort{\Self}{Self}

\defsort{\Agent}{Agent}
\defsort{\Entrant}{Entrant}
\defsort{\ActionType}{ActionType}
\defsort{\Moment}{Moment}
\defsort{\Boolean}{Formula}
\defsort{\PayOut}{PayOut}
\defsort{\Fluent}{Fluent}
\defsort{\Event}{Event}
\defsort{\Object}{Object}
\defsort{\RealTerm}{RealTerm}

\defsort{\Numeric}{Numeric}
\defsort{\Number}{Number}
\defsort{\Trolley}{Trolley}
\defsort{\Track}{Track}
\defsort{\Moveable}{Moveable}

\newcommand{\lsymbol}[1]{%
  \ensuremath{\mathit{#1}}}
\newcommand{\defsymbol}[2]{%
  \newcommand{#1}{\lsymbol{#2}}}
\defsymbol{\action}{action}
\defsymbol{\initially}{initially}
\defsymbol{\holds}{Holds}
\defsymbol{\happens}{happens}
\defsymbol{\clipped}{clipped}
\defsymbol{\initiates}{initiates}
\defsymbol{\terminates}{terminates}
\defsymbol{\prior}{prior}
\defsymbol{\interval}{interval}

\defsymbol{\does}{does}
\defsymbol{\plans}{plans}
\defsymbol{\act}{act}
\defsymbol{\react}{react}
\defsymbol{\payTot}{pay_{tot}}
\defsymbol{\fight}{fight}
\defsymbol{\coop}{coop}
\defsymbol{\enter}{enter}
\defsymbol{\stayout}{stayout}
\defsymbol{\learns}{learns}
\defsymbol{\payoff}{payoff}
\defsymbol{\position}{position}
\defsymbol{\dead}{dead}
\defsymbol{\damaged}{damaged}

\defsymbol{\onrails}{onrails}
\defsymbol{\switch}{switch}
\defsymbol{\drop}{drop}

\newcommand{\lconstant}[1]{%
  \ensuremath{\mbox{\textsf{#1}}}}
\newcommand{\defconstant}[2]{%
  \newcommand{#1}{\lconstant{#2}}}
\defconstant{\Enter}{Enter}
\defconstant{\StayOut}{StayOut}
\defconstant{\Fight}{Fight}
\defconstant{\Acquiesce}{Acquiesce}
\defconstant{\cs}{cs }
\newcommand{\trackA}{\ensuremath{track_1}}
\newcommand{\trackB}{\ensuremath{track_2}}

\newcommand{\lmodality}[1]{%
  \ensuremath{\mathbf{#1}}}
\newcommand{\defmodality}[2]{%
  \newcommand{#1}{\lmodality{#2}}}
\defmodality{\common}{C}
\defmodality{\knows}{K}
\defmodality{\believes}{B}
\defmodality{\perceives}{P}
\defmodality{\mental}{M}

\defmodality{\desires}{D}
\defmodality{\intends}{I}
\defmodality{\says}{S}
\defmodality{\ought}{O}

\newcommand{\lif}{\rightarrow}
\newcommand{\liff}{\leftrightarrow}
\newcommand{\sep}{\ \lvert \ }

\usepackage{booktabs}
\usepackage{mdframed}
\usepackage{paralist}

\usepackage{color}

\usepackage[usenames,dvipsnames,svgnames,table]{xcolor}

\newcommand{\DDE}{\ensuremath{\mathcal{{DDE}}}}
\newcommand{\DCEC}{\ensuremath{{\mathcal{{DCEC}}}}}

\newcommand{\DTE}{\ensuremath{\mathcal{{DTE}}}}
\newcommand{\type}[1]{\textsf{#1}}

\newcommand{\Intends}{\ensuremath{\mathbf{I}}}

\begin{document}
\date{March 2017}
\maketitle

\begin{abstract}

  The \textbf{doctrine of double effect} (\DDE) is a long-studied
  ethical principle that governs when actions that have both positive
  and negative effects are to be allowed.  The goal in this paper is
  to automate \DDE.  We briefly present \DDE, and use a first-order
  modal logic, the \textbf{deontic cognitive event calculus}, as our
  framework to formalize the doctrine.  We present formalizations of
  increasingly stronger versions of the principle, including what is
  known as the \textbf{doctrine of \emph{triple} effect}.  We then use
  our framework to successfully simulate scenarios that have been used
  to test for the presence of the principle in human subjects.  Our
  framework can be used in two different modes: One can use it to
  build \DDE-compliant autonomous systems from scratch; or one can use
  it to verify that a given AI system is \DDE-compliant, by applying a
  \DDE\ layer on an existing system or model.  For the latter mode,
  the underlying AI system can be built using any architecture
  (planners, deep neural networks, bayesian networks,
  knowledge-representation systems, or a hybrid); as long as the
  system exposes a few parameters in its model, such verification is
  possible.  The role of the \DDE\ layer here is akin to a (dynamic or
  static) software verifier that examines existing software modules.
  Finally, we end by sketching initial work on how one can apply our
  \DDE\ layer to the STRIPS-style planning model, and to a modified
  POMDP model. This is preliminary work to illustrate the feasibility
  of the second mode, and we hope that our initial sketches can be
  useful for other researchers in incorporating \DDE\ in their own
  frameworks.

\end{abstract}


\section{Introduction}
\label{sect:intro}
The \textbf{doctrine of double effect} (\DDE) is a long-studied
ethical principle that enables adjudication of ethically ``thorny''
situations in which actions that have both positive and negative
effects appear unavoidable for autonomous agents \cite{sep-dde}.  Such
situations are commonly called \textit{moral dilemmas}.  The simple
version of \DDE\ states that such actions, performed to ``escape''
such dilemmas, are allowed --- provided that
\begin{inparaenum}[1)] \item the harmful effects are not intended;
\item the harmful effects are not used to achieve the beneficial
  effects (harm is merely a \emph{side}-effect); and
\item benefits outweigh the harm by a significant
  amount. \end{inparaenum}
What distinguishes \DDE\ from, say, na\"{i}ve forms of
consequentialism in ethics (e.g.\ act utilitarianism, which holds that
an action is obligatory for an autonomous agent if and only if it
produces the most utility among all competing actions) is that purely
mental intentions in and of themselves, independent of consequences,
are considered crucial (as condition 2 immediately above conveys).  Of
course, every major ethical theory, not just consequentialism, has its
passionate proponents; cogent surveys of such theories make this plain
(e.g. see \cite{feldman_introductory_ethics}).  Even in machine
ethics, some AI researchers have explored not just consequentialism
and the second of the two dominant ethical theories, deontological
ethics (marked by an emphasis on fixed and inviolable principles said
by their defenders to hold no matter what the consequences of
abrogating them), but more exotic ones, for example contractualism
(e.g.\ see \cite{pereira_programming_machine_ethics}) and even
divine-command ethics (e.g.\ see \cite{sb_divine_command_robots}).
\DDE\ in a sense rises above philosophical debates about which
ethical theory is preferred.  The first reason is that empirical
studies have found that \DDE\ plays a prominent role in an ordinary
person's ethical decisions and judgments \cite{cushman2006role}.  For
example, in \cite{hauser2007dissociation}, a large number of
participants were asked to decide between action and inaction on a
series of moral dilemmas, and their choices adhered to \DDE,
irrespective of their ethical persuasions and backgrounds, and no
matter what the order in which the dilemmas were presented.  In
addition, in legal systems, criminality requires the presence of
malicious intentions \cite{fletcher1998basic}, and \DDE\ plays a
central role in many legal systems
\cite{allsopp2011doctrine,huxtable2004get}.\footnote{\newtext{On the
    surface, \emph{criminal negligence} might seem to require no
    intentions.  While that might be true, even in criminal negligence
    it seems rational to ask whether the negligence was accidental or
    something the ``suspect'' had control over. This suggests a milder
    form of intention, or something similar, but not exactly
    intention.}} Assuming that autonomous systems will be expected to
adjudicate moral dilemmas in human-like ways, and to justify such
adjudication, it seems desirable to seek science and engineering that
allows \DDE, indeed even nuanced, robust versions thereof, to be
quickly computed.





\section{Prior Work}
\label{sect:prior}
\newtext{We quickly review prior rigorous modeling of \DDE.  Mikhail
  in \cite{mikhail_moral_cognition_2011} presents one of the first
  careful treatments of the doctrine.  While the presentation of the
  doctrine makes use of some symbolism, the level of formalization is
  not amenable to automation.  \cite{bentzen2016principle} presents a
  model-theoretic formalization of a simple version of the doctrine.
  While this is an important first step, the calculus presented by
  Bentzen does not have any computational realization.  However, there
  are two independent strands of research with implementations for
  \DDE: that of Berreby et al.\ \shortcite{berreby2015modelling} and
  Pereira and Saptawijaya \shortcite{pereira2016counterfactuals}; both
  use logic programming.  Notably, while the Berreby et al.\
  explicitly eschew counterfactuals for modeling \DDE, Pereira and
  Saptawijaya model \DDE\ using counterfactuals.  To our knowledge,
  both the projects present one of the first formal models of
  \DDE\ that can be implemented. 

  It should be noted, however, that both of these formal systems are
  \textbf{extensional}, and it is well-known that when dealing with
  \textbf{intensional} states such as knowledge, belief, intention
  etc., extensional systems can quickly generate inconsistencies
  \cite{selmer_naveen_metaphil_web_intelligence} (see the appendix for
  more details).  The expressivity challenge is both quantificational
  and intensional; this challenge is acute for the logic-programming
  paradigm, as opposed to one based --- as is ours --- on formal
  languages beyond first-order logic and its variants, and proof
  theories beyond resolution and its derivatives.  In particular,
  \DDE\ requires elaborate structures for quantification (including,
  inevitably, first-order numerical quantifiers such as
  $\exists^k\!: k \in \mathbb{R}$, since quantification over utilities
  is essential), and many intensional operators that range over
  quantifiers, starting with the epistemic ones.  Needless to say,
  modeling and simulation at the propositional level, while truly
  excellent in the case of \cite{pereira2016counterfactuals}, is
  insufficiently expressive.

  Among the many empirical experiments centered around \DDE, the one
  in \cite{malle2015sacrifice} deserves a mention. Malle \textit{et al.} devise
  an experiment in which they place either a human or a robot as the
  central actor in a hypothetical \DDE\ scenario, and study an
  external viewer's moral judgement of action or inaction by the human
  or robot.  This study shows that humans view ethical situations
  differently when robots participate in such situations; and the
  study demonstrates the need for rigorous modeling of \DDE\ to build
  well-behaved autonomous systems that function in \DDE-relevant
  scenarios.}



\section{The Calculus}
\label{sect:calculus}

In this section, we present the \textbf{deontic cognitive event
  calculus} (\DCEC).  Dialects of this calculus have been used to
formalize and automate highly intensional reasoning processes, such as
the false-belief task
\cite{ArkoudasAndBringsjord2008Pricai} and \textit{akrasia} (succumbing to temptation to
violate moral principles) \cite{akratic_robots_ieee_n}.\footnote{Arkoudas and
  Bringsjord \shortcite{ArkoudasAndBringsjord2008Pricai} introduced
  the general family of \textbf{cognitive event calculi} to which
  \DCEC\ belongs.} \DCEC\ is a
sorted (i.e.\ typed) quantified modal logic (also known as sorted
first-order modal logic).  The calculus has a well-defined syntax and
proof calculus; see \cite{akratic_robots_ieee_n}.  The proof calculus
is based on natural deduction
\cite{gentzen_investigations_into_logical_deduction}, and includes all
the introduction and elimination rules for first-order logic, as well
as inference schemata for the modal operators and related structures.
A snippet of \DCEC\ is shown in the Appendix.

\subsection{Syntax}
\label{subsect:syntax}

\subsubsection{First-order Fragment} The first-order core of
\DCEC\ is the \emph{event calculus}
\cite{mueller_commonsense_reasoning}. Though we use the event
calculus, our approach is compatible with other calculi (e.g.\ the
\emph{situation calculus}) for modeling events and their effects.

\subsubsection{Modal Fragment} The modal operators present in the
calculus include the standard operators for knowledge $\knows$, belief
$\believes$, desire $\desires$, intention $\intends$, etc.  The general
format of an intensional operator is $\knows\left(a, t, \phi\right)$,
which says that agent $a$ knows at time $t$ the proposition $\phi$.
Here $\phi$ can in turn be any arbitrary formula.

The calculus also includes a dyadic deontic operator $\ought$.  The
unary ought in standard deontic logic is known to lead to
contradictions.  Our dyadic version of the operator blocks the
standard list of such contradictions, and beyond.\footnote{A nice
  version of the list is given lucidly in \cite{sep_deontic_logic}.}

\subsection{Semantics}

\subsubsection{First-order Fragment} The semantics for the
first-order fragment is the standard first-order semantics. The
truth-functional connectives $\land, \lor, \rightarrow, \lnot$ and
quantifiers $\forall, \exists$ for pure first-order formulae all have
the standard first-order semantics.\\

\subsubsection{Modal Fragment} The semantics of the modal operators
differs from what is available in the so-called
Belief-Desire-Intention (BDI) logics {\cite{bdi_krr_1999}} in many
important ways.  For example, \DCEC\ explicitly rejects possible-worlds
semantics and model-based reasoning, instead opting for a
\textit{proof-theoretic} semantics and the associated type of
reasoning commonly referred to as \textit{natural deduction}
\cite{gentzen_investigations_into_logical_deduction,proof-theoretic_semantics_for_nat_lang}.
Briefly, in this approach, meanings of modal operators are defined via
arbitrary computations over proofs, as we will soon see.

\subsection*{Reasoner (Theorem Prover)} Reasoning is performed through
a novel first-order modal logic theorem prover, \textsf{ShadowProver},
which uses a technique called \textbf{shadowing} to achieve speed
without sacrificing consistency in the system.  Extant first-order
modal logic theorem provers that can work with arbitrary inference
schemata are built upon first-order theorem provers.  They achieve the
reduction to first-order logic via two methods.  In the first method,
modal operators are simply represented by first-order predicates. This
approach is the fastest but can quickly lead to well-known
inconsistencies as demonstrated in
\cite{selmer_naveen_metaphil_web_intelligence}. In the second method,
the entire proof theory is implemented intricately in first-order
logic, and the reasoning is carried out within first-order logic.
Here, the first-order theorem prover simply functions as a declarative
programming system.  This approach, while accurate, can be
excruciatingly slow.  We use a different approach, in which we
alternate between calling a first-order theorem prover and applying
modal inference schemata.  When we call the first-order prover, all
modal atoms are converted into propositional atoms (i.e., shadowing),
to prevent substitution into modal contexts.  This approach achieves
speed without sacrificing consistency.  The prover also lets us add
arbitrary inference schemata to the calculus by using a
special-purpose language.  While we use the prover in our simulations,
describing the prover in more detail is out of scope for the present
paper.\footnote{The prover is available in both Java and Common Lisp
  and can be obtained at: \url{https://github.com/naveensundarg/prover}. The
  underlying first-order prover is SNARK available at:
  \url{http://www.ai.sri.com/~stickel/snark.html}.}

%
%

\section{Informal \DDE}
\label{sect:informal}
We now informally but rigorously present $\DDE$.  We assume we have at
hand an ethical hierarchy of actions as in the deontological case
(e.g.\ forbidden, neutral, obligatory); see \cite{bringsjord201721st}.
We also assume that we have a utility or goodness function for states
of the world or effects as in the consequentialist case.  For an
autonomous agent $a$, an action $\alpha$ in a situation $\sigma$ at
time $t$ is said to be $\DDE$-compliant \emph{iff}:

\begin{small}
\begin{enumerate}
\item[$\mathbf{C}_1$] the action is not forbidden (where we assume an
  ethical hierarchy such as the one given by Bringsjord
  \shortcite{bringsjord201721st}, and require that the action be
  neutral or above neutral in such a hierarchy);
\item[$\mathbf{C}_2$]  The net utility or goodness of the action is greater than some positive
  amount $\gamma$;
\item[$\mathbf{C}_{3a}$] the agent performing the action intends only the good effects;
\item[$\mathbf{C}_{3b}$] the agent does not intend any of the bad effects;
\item[$\mathbf{C}_4$] the bad effects are not used as a means to
  obtain the good effects; and
\item[$\mathbf{C}_5$] if there are bad effects, the agent would rather
  the situation be different and the agent not have to perform the
  action. That is, the action is unavoidable.
\end{enumerate}
\end{small}

See Clause 6 of Principle III in \cite{khatchadourian1988principle}
for a justification of of $\mathbf{C}_5$. This clause has not been
discussed in any prior rigorous treatments of $\DDE$, but we feel
$\mathbf{C}_5$ captures an important part of when $\DDE$ is normally
used, e.g.\ in unavoidable ethically thorny situations one would
rather not be present in. \newtext{$\mathbf{C}_5$ is necessary, as the
  condition is subjunctive/counterfactual in nature and hence may not
  always follow from $\mathbf{C}_1 - \mathbf{C}_4$, since there is no
  subjunctive content in those conditions.  Note that while
  \cite{pereira2016counterfactuals} model \DDE\ using counterfactuals,
  they use counterfactuals to model $\mathbf{C}_4$ rather than
  $\mathbf{C}_5$.}

That said, the formalization of $\mathbf{C}_5$ is quite difficult,
requiring the use of computationally hard counterfactual and
subjunctive reasoning.  We leave this aside here, reserved for future
work.


\section{Formal \DDE}
\label{sect:informal}
The formalization is straightforward given the machinery of \DCEC.
Let $\Gamma$ be a set of \textbf{background} axioms, which could
include whatever the given autonomous agent under consideration knows
about the world; e.g., its understanding of physics, knowledge and
beliefs about other agents and itself, etc.  The particular
\textbf{situation} that might be in play, e.g., \emph{``the autonomous
  agent is driving,''} is represented by a formula $\sigma$.  We use
ground fluents for effects.

We assume that we have a utility function $\mu$ that maps from fluents
and times to real-number utility values.  $\mu$ needs to be defined
only for ground fluents:


\begin{footnotesize}
\begin{equation*}
\begin{aligned}
\mu: \Fluent \times \Moment \rightarrow \mathbb{R}
\end{aligned}
\end{equation*}
\end{footnotesize}

\newtext{Good effects are fluents with positive utility, and bad
  effects are fluents that have negative utility.  Zero-utility
  fluents could be neutral fluents (which do not have a use at the
  moment).  }


\subsection{Defining \emph{means} $\rhd$}
The standard event calculus and \DCEC\ don't have any mechanism to say
when an effect is used as a \textbf{means} for another effect.  While
we could employ a first-order predicate and define axiomatically when
an effect is used as a means for another effect, we take a modal
approach that does not require any additional axioms beyond what is
needed for modeling a given situation.  Intuitively, we could say an
effect $e_1$ is a mere side effect for achieving another effect $e_2$
if by removing the entities involved in $e_1$ we can still achieve
$e_2$; otherwise we say $e_1$ is a means for $e_2$.  Our approach is
inspired by Pollock's \shortcite{Pollock1976-POLSR} treatment, and
while similarities can be found with the approach in
\cite{pereira2016counterfactuals}, we note that our definition
requires at least first-order logic.  Given a fluent $f$, we denote by
$\odot$ the set of all constants and function expressions in $f$.  For
example:


\begin{footnotesize}
\begin{equation*}
\begin{aligned}
&\odot\Big(\mathit{hungry}\big(\mathit{jack}\big)\Big) = \big\{\mathit{jack}\big\}\\
&\odot\Big(\mathit{married}\big(\mathit{jack,
  \mathit{sister}(mary)}\big)\Big) = \big\{\mathit{jack, sister(mary),
mary}\big\}
\end{aligned}
\end{equation*}
\end{footnotesize}


We need one more definition: the state of the world without a given
set of entities.  Let $\otimes(\Gamma, \theta)$, where $\Gamma$ is a
set of formulae and $\theta$ is a set of ground terms, be defined as
below:


\begin{footnotesize}
\begin{equation*}
\begin{aligned}
\otimes\Big(\Gamma, \theta
\Big) = \Big\{ \psi \in \Gamma \ \lvert\ \psi\mbox{
  does not
  contain any term in } \theta \Big\}
\end{aligned}
\end{equation*}
\end{footnotesize}

\newtext{Note that the above definition relies on the \textbf{Unique
    Names Assumption} commonly used in most formulations of the event
  calculus.  This assumption ensures that every object in the domain
  has at most one name or expression referring to it.  If this
  assumption does not hold, we can have the following slightly more
  complicated definition for $\otimes$.

\begin{footnotesize}
\begin{equation*}
\begin{aligned}
\otimes\Big(\Gamma, \theta
\Big) = \set{ \psi \in \Gamma   \,\middle|\ \begin{aligned}  \psi & \mbox{
  does not
  contain any  } s \mbox{ such that }\\ &   \exists t\in\theta: \Gamma\vdash s=t  \end{aligned}}
\end{aligned}
\end{equation*}
\end{footnotesize}
}

We introduce a new modal operator $\rhd$, \emph{means}, that says when
an effect is a means for another effect.


\begin{footnotesize}
\begin{equation*}
\begin{aligned}
\rhd: \Boolean
\times \Boolean \rightarrow \Boolean
\end{aligned}
\end{equation*}
\end{footnotesize}

\vspace{-0.1in}

The meaning of the operator is defined computationally below.  The
definition states that, given $\Gamma$, a fluent $f$ holding true at
$t_1$ causes or is used as a means for another fluent $g$ at time
$t_2$, with $t_2 > t_1$, \emph{iff} the truth condition for $g$
changes when we remove formulae that contain entities involved in $f$.
While this definition is far from perfect, it suffices as a first cut
  and lets us simulate experimental scenarios that have been used to
  test \DDE's presence in humans. 
  (Three other similar definitions hold when we look at combinations
  of fluents holding and not holding.)  The equation below follows
  (Note that $\vdash$ is non-monotonic, as it includes the event
  calculus):


\begin{footnotesize}
\begin{equation*}
\begin{aligned}
\Gamma \vdash \rhd\Big(\holds&\big(f,t_1\big), \holds\big(g,t_2\big)\Big)\\
&\mbox{\emph{iff} } 
\end{aligned}
\end{equation*}
\end{footnotesize}
\begin{footnotesize}
\begin{equation*}
\begin{aligned}
\left\{\begin{aligned}
\Gamma  &\vdash t_2 > t_1 { \wedge}\\
\left[\begin{aligned}\Gamma &\vdash \holds\big(f,t_1\big) { \land }\\
\Gamma &\vdash \holds\big(g,t_2\big)
\end{aligned}\right]&\Rightarrow 
\Big[\Gamma - \otimes\big(\Gamma,
\odot(f)\big)  \mathlarger{\vdash} \lnot \holds\big(g,t_2\big)\Big]
\end{aligned}\right\}
\end{aligned}
\end{equation*}
\end{footnotesize}


\newtext{For example, let $e_1$ be ``\emph{throwing a stone \textbf{s}
    at a window \textbf{w}}'' and $e_2$ be ``\emph{the window
    \textbf{w} getting broken}.''  We can see that $e_2$ is not just a
  mere side effect of $e_1$, and the definition works, since, if the
  stone is removed, $e_2$ wouldn't happen.  This definition is not
  perfect.  For instance, consider when there are common objects in
  both the events: the intuitiveness breaks down (but the definition
  still works).  We might for example let $e_1$ be ``\emph{hitting a
    window \textbf{w} with a bat \textbf{b}.}'' If the window and bat
  are not present, $e_2$ would not happen.}

\subsection{The Formalization}

\newtext{Note that the $\DDE\left(\Gamma, \sigma, a, \alpha, t,
  H\right)$ predicate defined below, though defined using \DCEC, lies
  outside of the formal language of \DCEC.  While \DDE\ is not fully
  formalized in \DCEC, the individual clauses
  $\mathbf{F}_1 - \mathbf{F}_4$ are.  This is how we can verify the
  conditions in the simulations described later.  It is trivial to
  define a new symbol and formalize the predicate in \DCEC: $\DDE1
  \Leftrightarrow \mathbf{F}_1 \land \mathbf{F}_2 \land \mathbf{F}_3
  \land \mathbf{F}_4$.

  What is not trivial, we concede, is how this works with other
  modalities.  For example, can we efficiently derive $\mathbf{K}(a,
  t_1, \mathbf{K}(b, t_2, \DDE\left(\Gamma, \sigma, a, \alpha, t,
  H\right)))$ given some other formulae $\Gamma$?  This could be
  difficult because the predicate's definition below involves
  provability, and one has to be careful when including a provability
  predicate.

  That said, for future work, we plan on incorporating this within an
  extended dialect of \DCEC.  One immediate drawback is that while we
  can have a \emph{system}-level view of whether an action is
  \DDE-sanctioned, agents themselves might not know that.  For
  example, we would like to able to write down \textit{``a knows that
    b knows that c's action is \DDE-sanctioned.''}}

Given the machinery defined above, we now proceed to the
formalization.  Assume, for any action type $\alpha$ carried out by an
agent $a$ at time $t$, that it initiates the set of fluents
$\alpha_I^{a,t}$, and terminates the set of fluents $\alpha_T^{a,t}$.
Then, for any action $\alpha$ taken by an autonomous agent $a$ at time
$t$ with background information $\Gamma$ in situation $\sigma$, the
action adheres to the doctrine of double effect up to a given time
horizon $H$, that is $\DDE\left(\Gamma, \sigma, a, \alpha, t,
H\right)$ \emph{iff} the conditions below hold:

\begin{small}

\begin{mdframed}[linecolor=white, frametitle= Formal Conditions for \DDE, frametitlebackgroundcolor=gray!25, backgroundcolor=gray!10, roundcorner=8pt]
\begin{enumerate}
\item[$\mathbf{F_1}$] $\alpha$ carried out at $t$ is not forbidden. That is:
\begin{footnotesize}
\begin{equation*}
\begin{aligned}
&\Gamma \not\vdash \lnot \ought\Big(a, t, \sigma, \lnot \happens\big(action(a, \alpha),t\big)\Big)
\end{aligned}
\end{equation*}
\end{footnotesize}

\item[$\mathbf{F_2}$] The net utility is greater than a given positive
  real $\gamma$:

\begin{footnotesize}
\begin{equation*}
\begin{aligned}
    \Gamma \vdash \mathlarger{\mathlarger{‎‎\sum}}_{y=t+1}^{H‎}\Bigg(\sum_{f\in{\alpha_{I}^{a,t}}} \mu(f, y) - \sum_{f\in{\alpha_{T}^{a,t}}} \mu(f, y)\Bigg) > \gamma
\end{aligned}
\end{equation*}
\end{footnotesize}

\item[$\mathbf{F_{3a}}$] The agent $a$ intends at least one good
  effect.  ($\mathbf{F}_2$ should still hold after removing all
  other good effects.)  There is at least one
  fluent $f_g$ in $\alpha_{I}^{a,t}$ with
  $\mu\left(f_g, y\right) > 0$, or $f_b$ in $\alpha_{T}^{a,t}$ with
  $\mu\left(f_b, y\right) < 0$, and some $y$ with $t<y\leq H$ such
  that the following holds:
\begin{footnotesize}
\begin{equation*}
\begin{aligned}
\Gamma &\vdash \left(\begin{aligned} &\exists f_g \in \alpha_{I}^{a,t}
    \ 
    \Intends\Big(a, t, 
    \holds\big(f_g,y\big)\Big)\\ & \ \ \ \ \ \ \ \ \ \ \  \ \ \ \ \ \
    \ \ \  \ \ \ \lor\\
 &\exists f_b\in \alpha_{T}^{a,t} \ \Intends\Big(a, t, 
    \lnot \holds\big(f_b, y\big)\Big) 
\end{aligned}\right)
\end{aligned}
\end{equation*}
\end{footnotesize}

\item[$\mathbf{F_{3b}}$] The agent $a$ does not intend any bad effect.
  For all fluents $f_b$ in $\alpha_{I}^{a,t}$ with
  $\mu\left(f_b, y\right) < 0$, or $f_g$ in $\alpha_{T}^{a,t}$ with
  $\mu\left(f_g, y\right) > 0$, and for all $y$ such that $t<y\leq H$
  the following holds: 
\begin{footnotesize}
\begin{equation*}
\begin{aligned}
\Gamma \not&\vdash \Intends\Big(a, t,
    \holds\big(f_b,y\big)\Big) \mbox{ and }\\
\Gamma \not&\vdash \Intends\Big(a, t,
    \lnot \holds\big(f_g,y\big)\Big) 
\end{aligned}
\end{equation*}
\end{footnotesize}


\item[$\mathbf{F_{4}}$] The harmful effects don't cause the good
  effects.  Four permutations, paralleling the definition of $\rhd$
  above, hold here.  One such permutation is shown below.  For any bad
  fluent $f_b$ holding at $t_1$, and any good fluent $f_g$ holding at
  some $t_2$, such that $t< t_1, t_2\leq H$, the following holds:

 \begin{footnotesize} \begin{equation*} \begin{aligned}
        \Gamma \vdash \lnot \rhd\Big(\holds\big(f_b, t_1\big), \holds
        \big(f_g, t_2\big)\Big) \end{aligned} \end{equation*}
  \end{footnotesize}


\item[$\mathbf{F_{5}}$] This clause requires subjunctive reasoning.
  The current formalization ignores this stronger clause.  There has
  been some work in computational subjunctive reasoning that we hope
  to use in the future; see \cite{Pollock1976-POLSR}.

\end{enumerate}
\end{mdframed}

\end{small}

\vspace{0.1in}

\subsubsection{Doctrine of Triple Effect}  The doctrine of triple
effect (\DTE) was proposed in \cite{kamm2007intricate} to account for
scenarios where actions that are viewed as permissible by most
philosophers and deemed as such by empirical studies (e.g.\ the switch
action in the third scenario in \cite{hauser2007dissociation}) are not
sanctioned by \DDE, as they involve harm being used as a means to
achieve an action.  \DTE\ allows such actions as long as the harm is
not explicitly intended by the agent.  Note that our version of
\DDE\ subsumes \DTE\ through condition $\mathbf{C}_4$. 


\section{ Scenarios}
\label{sect:scenarios}
The trolley problems are quite popular in both philosophical and
empirical studies in ethics.  Hauser \textit{et al.}
\shortcite{hauser2007dissociation} found empirical support that
\DDE\ is used by humans, courtesy of experiments based on a set of
trolley problems.  They use a set of 19 trolley problems in their
experimentation, and describe in detail four of these.  We consider
the first two of these problems in our study here.  The problem
scenarios are briefly summarized below; common to both this setup:
There are two tracks $\trackA$ and $\trackB$.  There is a trolley
loose on $\trackA$ heading toward two people $P_1$ and $P_2$ on
$\trackA$; neither person can move in time.  If the trolley hits them,
they die.  The goal is to save this pair.\footnote{For computational
  purposes, the exact number of persons is not important as long as it
  is greater than one.}

\begin{enumerate}

\item[\textbf{Scenario 1}] There is a switch that can route the
  trolley to $\trackB$.  There is a person $P_3$ on $\trackB$.
  Switching the trolley to $\trackB$ will kill $P_3$.  Is it okay to
  switch the trolley to $\trackB$?

\item[\textbf{Scenario 2}] There is no switch now, but we can push
  $P_3$ onto the track in front of the trolley.  This action will
  damage the trolley and stop it; it will also kill $P_3$.  Is it okay
  to push $P_3$ onto the track?
\end{enumerate}
\DDE-based analysis tells us it is okay to switch the trolley in
\textbf{Scenario 1}, as we are killing the person merely as a side
effect of saving $P_1$ and $P_2$.  In \textbf{Scenario 2}, similar
analysis tells us it is not okay to push $P_3$, because we are using
that person as a means toward our goal.


\section{Simulations}
\label{sect:simulations}
At the core of our simulation is a formalization of the basic trolley
scenario based on the event calculus.  We use a discrete version of
the event calculus, in which time is discrete, but other quantities
and measures, such as the utility function, can be continuous.  We
have the following additional sorts: \Trolley\ and \Track.  We also
declare that the \Agent\ and the \Trolley\ sorts are subsorts of a
\Moveable\ sort, the instances of which are objects that can be placed
on tracks and moved.  We use the following additional core symbols:

\vspace{-0.03in}
\begin{footnotesize}
\begin{equation*}
\begin{aligned}
&\position: \Moveable \times \Track \times \Number \rightarrow \Fluent
\\
&\dead: \Agent \rightarrow \Fluent \\
&\onrails: \Trolley \times \Track \rightarrow \Fluent\\
&\switch: \Trolley \times \Track \times \Track \rightarrow \ActionType\\
&\mathit{push}: \Agent \times \Track \times \Number \rightarrow \ActionType\\
\end{aligned}
\end{equation*}
\end{footnotesize}

\vspace{-0.05in}

The utility
function $\mu$ is defined as follows:

\begin{footnotesize}
\begin{equation*}
\begin{aligned}
    \mu(f,t) =
    \begin{cases*}
      -1 & if $f \equiv \dead(P)$ \\
      0        & otherwise
    \end{cases*}
\end{aligned}
\end{equation*}
\end{footnotesize}
\vspace{-0.05in}

We set the threshold $\gamma$ at $0.5$.  The simulation starts at time
$t=0$ with the only trolley, denoted by $\mathit{trolley}$, on
\trackA.  We have an event-calculus trajectory axiom shown below as
part of $\Gamma$:

\begin{footnotesize}
\begin{equation*}
\begin{aligned}
\forall t:\Trolley &, track:\Track, s:\Moment \\
&\Big[\mathit{Trajectory}\Big(\onrails(t, track), s, \position(t,
track, \Delta), \Delta \Big)\Big]
\end{aligned}
\end{equation*}
\end{footnotesize}

The above axiom gives us the trolley's position at different points of
time.  $\Gamma$ also includes axioms that account for non effects.
For example, in the absence of any actions, we can derive:

\begin{footnotesize}
\begin{equation*}
\begin{aligned}
\Gamma \vdash \holds\Big(\position\big(\mathit{trolley}, \trackA, 23\big), 23\Big)
\end{aligned}
\end{equation*}
\end{footnotesize}

We also have in the background $\Gamma$ a formula stating that in the
given trolley scenario the agent ought to save both $P_1$ and $P_2$.
Ideally, while we would like the agent to arrive at this obligation
from a more primitive set of premises, this setup is closer to
experiments with human subjects in which they are asked explicitly to
save the persons.  Note the agent performing the action is simply
denoted by $I$, and let the time of the test be denoted by $now$.

\begin{footnotesize}
\begin{equation*}
\begin{aligned}
\ought\left(I, now,
\sigma_{trolley}, 
\left[\begin{aligned}
&\lnot \exists t: \Moment\ \holds\left(\dead(P_1,t)\right) \land \\
&\lnot \exists t: \Moment\ \holds\left(\dead(P_2,t)\right)
\end{aligned}\right]
\right)
\end{aligned}
\end{equation*}
\end{footnotesize}

Given that the agent knows that it is now in situation
$\sigma_{trolley}$, and the agent believes that it has the above
obligation, we can derive from \DCEC's inference schemata what the
agent intends:
\begin{footnotesize}
\begin{equation*}
\begin{aligned}
&\left\{
\begin{aligned}
  &\knows\Big(I, now, \sigma_{trolley}\Big),\\
  &\believes\left(I, now, \ought\left(
      \begin{aligned} & I, now, \sigma_{trolley}, \\
        &\left[\begin{aligned}
            \lnot \exists t: \Moment\ & \holds\Big(\dead\big(P_1,t\big)\Big) 
            \\ &\land \\
            \lnot \exists t: \Moment\ &\holds\Big(\dead\big(P_2,t\big)\Big)
          \end{aligned}\right]
      \end{aligned}\right)\right),\\
  &\ought\Bigg(I, now,
  \sigma_{trolley}, 
  \left[\begin{aligned}
      &\lnot \exists t: \Moment\ \holds\left(\dead(P_1,t)\right) \land \\
      &\lnot \exists t: \Moment\ \holds\left(\dead(P_2,t)\right)
    \end{aligned}\right]
  \Bigg)
\end{aligned}\right\}\\
&\ \ \ \ \ \ \mathlarger{\mathlarger{\vdash}}  \ \ \
\Intends\left(I, now,
\left[\begin{aligned}
&\lnot \exists t: \Moment\ \holds\Big(\dead\big(P_1,t\big)\Big) \land \\
&\lnot \exists t: \Moment\ \holds\Big(\dead\big(P_2,t\big)\Big)
\end{aligned}\right]
\right)
\end{aligned}
\end{equation*}
\end{footnotesize}

In both the simulations, $P_1$ is at position $4$ and $P_2$ is at
position $5$ on \trackA.  In \textbf{Scenario 1}, $P_3$ is at position $3$
on \trackB, and the train can be switched from position $3$ on
\trackA\ to position $0$ on \trackB. 

In \textbf{Scenario 2}, we push $P_3$ onto position 3 on \trackA.  The
total number of formulae and run times for simulating the two
scenarios with and without the actions are shown below.  Note these
are merely event-calculus simulation times.  These are then used in
computing $\DDE(\Gamma,\sigma, a, \alpha, t, H)$.  The event-calculus
simulation helps us compute $\mathbf{F}_2$.


\begin{footnotesize}
\begin{tabular}{lccc}  
\toprule
& & \multicolumn{2}{c}{\textbf{Simulation Time} (s) }\\
\cmidrule(r){3-4}
\textbf{Scenario}   & ${\vert\Gamma\vert}$ & \emph{No action} & \emph{Action performed} \\
\midrule
\textbf{Scenario 1} &39 & 0.591   & 1.116     \\
\textbf{Scenario 2 } &38  & 0.602   & 0.801  \\
\bottomrule
\end{tabular}
\end{footnotesize}

\vspace{0.15in}

\textsf{ShadowProver} was then used to verify that $\mathbf{F}_1$,
$\mathbf{F}_{3a}$, and $\mathbf{F}_{3b}$ hold.  Both the scenarios
combined take $0.57$ seconds for $\mathbf{F}_1$, $\mathbf{F}_{3a}$,
and $\mathbf{F}_{3b}$.  The scenarios differ only in $\mathbf{F}_4$.
The pushing action fails to be \DDE-compliant due to
$\mathbf{F}_4$. For verifying that $\mathbf{F}_4$ holds in
\textbf{Scenario 1} and doesn't hold in \textbf{Scenario 2}, it takes
$0.49$ seconds and $0.055$ seconds, respectively.\footnote{All the
  axioms for the two simulations, \textsf{ShadowProver}, and the
  combined $\DDE$ implementation can be obtained here:
  \url{https://goo.gl/9KU2L9}.}


\section{On Operationalizing the Principle}
\label{sect:operationalize}
Given the above formalization, it's quite straightforward to build
logic-based systems that are \DDE-compliant.\footnote{For examples of
  logic-based systems in pure first-order logic, see
  \cite{mueller_commonsense_reasoning}.}
But how do we apply the above formalization to existing models and
systems that are not explicitly logic-based?  We lay down a set of
conditions such models must satisfy to be able to verify that they are
\DDE-compliant.  We then sketch how we could use \DDE\ in two such
modified systems: a STRIPS-like planner and a POMDP type model.

The problem now before us is: Given a system and a utility function,
can we say that the system is \DDE-compliant?  No, we need more
information from the system.  For example, we can have two systems in
the same situation, the same utility functions and same set of
available actions.\footnote{Where does a utility function come from?
  The obvious way to get a utility function seems to be to learn such
  a function.  There are good arguments that such learning can be very
  hard \cite{arnold2017value}.  For now, we are not concerned with how
  such a utility function is given to us.  For exposition and economy
  assume that it already exists. } One system can be \DDE-compliant
while the other is not.  For example, assume that we have two
autonomous driving systems $d_1$ and $d_2$.  Assume that $d_2$ has
learned to like killing dogs and intends to do so if possible during
its normal course of operation.  While driving, both come across a
situation where the system has to hit either a human or a dog.  In
this scenario, $d_1$'s action to hit the dog would be \DDE-compliant
while $d_2$'s action will not be.  Therefore, the formalization
requires that we have access to an agent's intentions at all times.

\newtext{One common objection to requiring that intentions be separate
  from utilities states that utilities can be used to derive
  intentions.  This is mistaken: it is not always possible to derive
  intentions from a utility function.  For example, there might be a
  state that has high utility but the agent might not intend to
  realize that state, as it could be out of reach for that agent (low
  perceived probability of success).

  For instance, winning a million dollars ($w$) has high utility, but
  most rational agents might not intend $w$, as they know this event
  is (alas) out of their reach.  This holds for similar high-utility
  states.

  At a minimum, we believe utility and perceived probability of
  success go into an agent's intentions.  This seems to align with the
  human case when we are looking at motivations,
  i.e.\ \textit{expectancy-value} theory.  How motivations could
  transform into intentions is another open research question.}


\subsection{Requirements}

Practically speaking, there is a spectrum of systems that our
techniques will be dealing with.  At one end, we will encounter
systems that are complete black boxes taking in percepts from the
environment and spitting out actions.  Since \DDE\ requires us to look
at intentions of systems, such black-box systems will be impossible to
verify.  We can of course ask the system to output its intention
through language as one of its possible actions, but this means that
we are relying on the system's honest reporting of its internal
states.  At the other end of the spectrum, we have complete white-box
systems.  We can be fully confident that we can get what the system
intends, believes, knows, etc.\ at any point in time.  Verifying such
systems is possible, in theory at least.  While we don't know what
kind of shape autonomous systems will take and where they will fall in
the spectrum, we can explicitly list information we need from such
systems before we can start the verification process.  One such
specification follows.

\begin{footnotesize}
\begin{mdframed}[linecolor=white, frametitle= Gray Box Requirement, frametitlebackgroundcolor=gray!25, backgroundcolor=gray!10, roundcorner=8pt]
 Given any autonomous system $a$, at any point of time $t$, we should
 \emph{at least} be able to assert the following, if true, in order to
 verify that it is $\DDE$-compliant:

\begin{enumerate}
  \item The system's intentions: $(\lnot)\Intends\left(a, t, \phi\right)$
  \item Prohibitions: $\lnot\ought\left(a, t, \sigma, \lnot\phi\right)$
\end{enumerate}
\end{mdframed}
\end{footnotesize}

How would we go about applying the formalization to other formal
systems?  We very briefly sketch two examples.

\subsubsection{STRIP-like Planner} We first look at a STRIPS-style
planning system.  Briefly, a STRIPS-style planner has a set of actions
$\{a_i\}$ and a set of states $\{s_i\}$.  The states are nothing but
sets of formulae or atoms.  The individual formulae would be our
effects.  Each action $a$ has a set of preconditions $pre(a)$, a set
of formulae that should hold in a given state to execute that action
in that state.  After executing an action $a$ in a state $s$, the new
state is given by $s \cup \mathit{additions}(a) - \mathit{deletions}(a)$.
The planner is given an explicit goal $\phi$.  This means that we know
$(\lnot)\intends(a, t, \phi)$ trivially.  If we have an ethical
hierarchy for the available set of actions, we then satisfy the
gray-box requirement.  What is then needed is a definition for $\rhd$,
an effect used as means for another effect.  The formalism gives us
one possible way to define $\rhd$.  A plan $\rho$ is nothing but a
sequence of actions.  Given a plan $\rho$, we say an effect $e_1$ is
used as means for another effect $e_2$, if $e_1 \in pre(a_1)$, $a_1$
is an action in the plan and $e_2 \in \mathit{additions}(a_2)$, and
$a_1$ comes before $a_2$.

\subsubsection{POMDP-derived System} Partially observable Markov
decision process (POMDP) models have been quite successful in a large
number of domains.  It is highly likely that some of the first
autonomous systems might be based on POMDPs.  We note that in such
models, the only goal is to maximize a reward function.  Another issue
is that states are atomic.  In order to discern between good and bad
effects, we would need states to be decomposed into smaller
components.  One possible approach could use \textit{factored markov
  decision processes}, which are MDPs in which states are represented
as a mapping $m$ from a set of state variables $\Theta =\{s_1, s_2,
\ldots, s_n\}$ to a set of values $\mathcal{V}$.  Here the utility and
reward function could be defined on the assignments; i.e., $reward(s)
= \sum \mu(s_i \leadsto \nu)$, where $\mu$ assigns a utility value to
a particular assignment of a state variable.  Additionaly, the
formalism could specify one or more goal states that the model seeks
to attain while maximizing the reward along the way, giving us
$(\lnot)\intends(a,t,\phi)$.



\section{Heirarchies of Doctrines}
\label{sect:strength}

Our formalization, summarized in the equation below, gives rise to
multiple hierarchies of the doctrine.  We discuss some of the
hierarchies below.
\begin{footnotesize}
\begin{equation*}
\begin{aligned}
\DDE\left(\Gamma, \sigma, a, \alpha, t, H\right)
\Leftrightarrow \mathbf{F}_1 \land \mathbf{F}_2\land
\mathbf{F}_3\land \mathbf{F}_4
\end{aligned}
\end{equation*}
\end{footnotesize}
\vspace{-0.2in}

\begin{small}
\begin{enumerate}
\item[\textbf{Horizon}] One obvious knob in the above equation is the
  horizon $H$.  Increasing $H$ will give us stronger versions of the
  doctrine.  Since our formalization is in first-order modal logic,
  the horizon need not be finite: we could set the horizon to
  infinity, $H=\omega$, and still obtain a tractable model, as long as
  we carefully develop our formalization.\footnote{It's a well-known
    fundamental result that first-order logic can handle infinite
    models with a finite number of axioms; see e.g.\ Ch.\ 12 in
    \cite{boolos_jeffrey_5thed}.}

\item[\textbf{Agent Generality}] Instead of just checking whether an
  action at a given time is \DDE-compliant, we could ask whether an
  autonomous agent $a$ in a given situation $\sigma$ will be
  \DDE-compliant at all times. This gives us the following
  condition:
\begin{footnotesize}
\begin{equation*}
\begin{aligned}
\forall \alpha\!:\!\ActionType,t\!:\!\Moment.\ \DDE\left(\Gamma, \sigma, a, \alpha, t, H\right)
\end{aligned}
\end{equation*}
\end{footnotesize}
\vspace{-0.2in}

\item[\textbf{Situation Generality}] In the hierarchy above, the
  quantification was over objects.  We could ask whether an autonomous
  agent would be \DDE-compliant in all situations.  That would
  correspond to a quantification over formulae (see centered formula
  immediately below), something not supported in the version of \DCEC\
  used herein.
\begin{footnotesize}
\begin{equation*}
\begin{aligned}
\forall \sigma\!:\!\Boolean,\alpha\!:\!\ActionType,t\!:\!\Moment.\DDE\left(\Gamma, \sigma, a, \alpha, t, H\right)
\end{aligned}
\end{equation*}
\end{footnotesize}
\vspace{-0.2in}

\item[\textbf{Counterfactual Reasoning}] The presence or absence of
  counterfactual reasoning in $\mathbf{F}_5$ would correspond to a
  very strong version of the doctrine, but one that would also be very
  hard to automate in the general case.  We note that there are
  hierarchies of counterfactual reasoning (see
  \cite{Pollock1976-POLSR}) that could correspond to hierarchies of
  versions of \DDE.

\end{enumerate}
\end{small}


\section{Conclusion}
\label{sect:conclusion}
We now quickly summarize the chief contributions of the foregoing, and
end by presenting future lines of work.  Our primary contribution is
the presentation of a novel computational logic, or cognitive
calculus, in which important versions of \DDE\ are formalized.  As a
part of this calculus, we formalized an effect being used as a means
for another effect via the modal operator $\rhd$.  We also supplied an
informal but rigorous version, $\mathbf{C}_1-\mathbf{C}_4$, of the
doctrine itself, from which we built our formalization
$\mathbf{F}_1-\mathbf{F}_4$.  Included in this formalization is the
clause $\mathbf{C_5}/\mathbf{F}_5$, which requires subjunctive and
counterfactual reasoning, an aspect that hitherto has simply not been
considered in any systematic treatment of \DDE.  Our formalization
subsumes the doctrine of \emph{triple} effect, \DTE; we have achieved
the first computational simulations of the doctrine.  A byproduct of
these simulations is an event-calculus formalization of a demanding
class of trolley problems (widely used in empirical and philosophical
studies of ethics).  We noted that our formalization gives rise to
hierarchies of doctrines with varying strengths.  Our readers can
choose a particular strength doctrine that fits their needs.

Future work includes simulating more intricate ``ethically thorny''
scenarios.  Despite our progress, we note that our formalization is
devoid of any mechanisms for handling uncertainty, and we are in the
process of extending our work to include reasoning based on
probabilistic versions of \DCEC.\footnote{There exist probabilistic
  versions of the event calculus.
  \IGNORE{\cite{Skarlatidis:2015:PEC:2737801.2699916}.} We will
  leverage similar work.}  We also note that we have not said much
about how our formalization could interact with an autonomous learning
agent.  We observe that even the possibility that such an intricate
principle as \DDE/\DTE\ is learnable using existing learning
frameworks remains open to question \cite{arnold2017value}.  In the
short term, a guaranteed-to-be-fruitful but less ambitious area of
development will be the deployment of our mechanization of \DDE\ in
existing systems, and adapting existing formal models, as briefly
discussed above, to exploit this mechanization.  Finally, we note that
since we are using first-order (multi) modal logic, we will eventually
run into efficiency issues, as even vanilla first-order logic's
decision problem, $\Gamma \vdash \gamma$, is Turing-undecidable.
There are a number of techniques to mitigate this issue.  One approach
is to exploit a library of commonly used proof patterns codified in a
denotational proof language; see
\cite{ArkoudasAndBringsjord2008Pricai}.  We are cautiously optimistic,
as many formal enterprises outside of AI (e.g.\ software verification
\cite{khasidashvili2009verifying} and formal physics
\cite{stannett2014using}) routinely face such challenges and surmount
them.


\section*{Acknowledgements}
\label{sect:ack}

We are grateful to the Office of Naval Research for funding that
enabled the research presented in this paper.  We also thank Dr.\
Daniel Thero for reading a draft of the paper and providing valuable
feedback.  We are also grateful for the insightful reviews provided by
the five anonymous referees.
\appendix


\newpage
\clearpage

\section{Deontic Cognitive Event Calculus}
\label{sect:appendixA}

We provide here a short primer on the deontic cognitive event calculus
(\DCEC).  A calculus is a set of axioms in a formal logic.  For
example, the event calculus is a set of axioms couched in first-order
logic.
\DCEC\ is a set of axioms in sorted first-order modal logic (also known
as sorted quantified modal logic) that subsumes the event calculus.

While first-order logic is an \textbf{extensional} system, modal
logics are \textbf{intensional} systems.  Note that there is a
profound difference between intension vs.\ intention.  One can have an
intention to bring something about; this is traditionally captured by
particular intensional operators.  In other words, put concretely, the
intention operator $\mathbf{I}$ is an intensional operator, but so is
$\mathbf{D}$ for desire, $\mathbf{B}$ for believes, and $\mathbf{P}$
for perceives, etc.

\DCEC\ is \textbf{intensional} in the sense that it includes
intensional operators.  Unfortunately, the situation is further
confused by the fact that traditionally in philosophy of mind,
intentionality means the so-called ``aboutness'' of some mental
states, so that my belief that Melbourne is beautiful is in this sense
intentional, while my mental state has nothing to do with intending
something.  Most logicians working in formal intensional systems
believe that at least intensional logic is required to formalize
intentional states \cite{zalta1988intensional}. One simple reason is that using plain first-order logic leads to
unsound inferences as shown below.  In the inference below, we have an
agent $a$ that knows that the killer in a particular situation is the
person that owns the knife.  Agent $a$ does not know that the
$\mathit{Moe}$ is the killer, but it's true that $\mathit{Moe}$ is the
owner of the knife.  If the knowledge operator $\mathbf{K}$ is a
simple first-order predicate, we will get the proof shown below,
which produces a contradiction from sound premises. See
\cite{selmer_naveen_metaphil_web_intelligence} for a sequence of
stronger representation schemes in first-order logic for knowledge and
belief that still result in inconsistencies.

\begin{scriptsize}
\begin{mdframed}[frametitle=Modeling Knowlege (or any Intension) in First-order
  Logic ,
  frametitlebackgroundcolor=gray!25,linecolor=white,backgroundcolor=gray!10]

\begin{scriptsize}
\begin{equation*}
\begin{aligned}
&\fbox{1}\ \ \mathbf{K}\left(a,\
  \mathsf{Killer}\left(\mathit{owner}\left(\mathit{knife}\right)\right)\right) \mbox{
  {\color{gray}; given}} \\
&\fbox{2}\ \ \lnot \mathbf{K}\left(a,\mathsf{Killer}\left(\mathit{Moe}\right)\right) \mbox{
  {\color{gray}; given}}\\
&\fbox{3}\ \ \mathit{Moe} = \mathit{owner}\left(\mathit{knife}\right)  \mbox{
  {\color{gray}; given}}\\
&\fbox{4}\ \ \mathbf{K}\left(a,\mathsf{Killer}\left(\mathit{Moe}\right)\right)  \mbox{
  {\color{gray}; first-order inference from \fbox{3} and \fbox{1}}}\\
& \fbox{5}\ \ \mathbf{\bot}  \mbox{
  {\color{gray}; first-order inference from \fbox{4} and \fbox{2}}}
\end{aligned}
\end{equation*}
\end{scriptsize}
\end{mdframed}
\end{scriptsize}

\subsection{Syntax of Deontic Cognitive Event Calculus}
\DCEC\ is a sorted calculus.  A sorted system can be thought of as
being analogous to a typed single-inheritance programming language.
We show below some of the important sorts used in \DCEC.  Among these,
the
\type{Agent}, \type{Action} and \type{ActionType} sorts are not native
to the event calculus.  
\vspace{0.08in}

\begin{scriptsize}
\begin{tabular}{lp{5.8cm}}  
\toprule
Sort    & Description \\
\midrule
\type{Agent} & Human and non-human actors.  \\

\type{Time} &  The \type{Time} type stands for
time in the domain.  E.g.\ simple, such as $t_i$, or complex, such as
$birthday(son(jack))$. \\

 \type{Event} & Used for events in the domain. \\
 \type{ActionType} & Action types are abstract actions.  They are
  instantiated at particular times by actors.  Example: eating.\\
 \type{Action} & A subtype of \type{Event} for events that occur
  as actions by agents. \\
 \type{Fluent} & Used for representing states of the world in the
  event calculus. \\
\bottomrule
\end{tabular}
\end{scriptsize}
\vspace{0.08in}

The figures below show the syntax and inference schemata of \DCEC.
The syntax is quantified modal logic.  Commonly used function and
relation symbols of the event calculus are included.  Particularly,
note the following modal operators: $\mathbf{P}$ for perceiving a
state, $\mathbf{K}$ for knowledge, $\mathbf{B}$ for belief,
$\mathbf{C}$ for common knowledge, $\mathbf{S}$ for agent-to-agent
communication and public announcements, $\mathbf{B}$ for belief,
$\mathbf{D}$ for desire, $\mathbf{I}$ for intention, and finally and
crucially, a dyadic deontic operator $\mathbf{O}$ that states when an
action is obligatory or forbidden for agents. It should be noted that
\DCEC\ is one specimen in a \emph{family} of easily extensible
cognitive calculi.  Since the semantics of \DCEC\ is proof-theoretic,
as long as a new construct has appropriate inference schemata, the
extension is sanctioned.

\vspace{0.08in}
\begin{scriptsize}
\begin{mdframed}[linecolor=white, frametitle=Syntax, frametitlebackgroundcolor=gray!25, backgroundcolor=gray!10, roundcorner=8pt]
\begin{equation*}
 \begin{aligned}
    \mathit{S} &::= 
    \begin{aligned}
      & \Object \sep \Agent \sep \ActionType \sep \Action \sqsubseteq
      \Event \sep \Moment \sep \Boolean \sep \Fluent \\
    \end{aligned} 
    \\ 
    \mathit{f} &::= \left\{
    \begin{aligned}
      & \action: \Agent \times \ActionType \rightarrow \Action \\
      &  \initially: \Fluent \rightarrow \Boolean\\
      &  \holds: \Fluent \times \Moment \rightarrow \Boolean \\
      & \happens: \Event \times \Moment \rightarrow \Boolean \\
      & \clipped: \Moment \times \Fluent \times \Moment \rightarrow \Boolean \\
      & \initiates: \Event \times \Fluent \times \Moment \rightarrow \Boolean\\
      & \terminates: \Event \times \Fluent \times \Moment \rightarrow \Boolean \\
      & \prior: \Moment \times \Moment \rightarrow \Boolean\\
    \end{aligned}\right.\\
        \mathit{t} &::=
    \begin{aligned}
      \mathit{x : S} \sep \mathit{c : S} \sep f(t_1,\ldots,t_n)
    \end{aligned}
    \\ 
    \mathit{\phi}&::= \left\{ 
    \begin{aligned}
     & t:\Boolean \sep  \neg \phi \sep \phi \land \psi \sep \phi \lor
     \psi \sep \perceives (a,t,\phi)  \sep \knows(a,t,\phi) \sep
     \common(t,\phi) \\ &
 \says(a,b,t,\phi) 
     \sep \says(a,t,\phi) \sep  \believes(a,t,\phi) \sep \desires(a,t,\holds(f,t')) \sep
     \intends(a,t,\phi)\\
     & \ought(a,t,\phi,(\lnot)\happens(action(a^\ast,\alpha),t'))
      \end{aligned}\right.
  \end{aligned}
\end{equation*}
\end{mdframed}
\end{scriptsize}

The figure below shows the inference schemata for
\DCEC.  $R_\mathbf{K}$ and $R_\mathbf{B}$ are inference schemata that
let us model idealized agents that have their knowledge and belief
closed under the \DCEC\ proof theory.  While normal humans are not
dedcutively closed, this lets us model more closely how deliberate
agents such as organizations and more strategic actors reason. (Some
dialects of  cognitive calculi restrict the number of iterations on intensional
operators.)  $R_1$ and $R_2$ state respectively that it is common
knowledge that perception leads to knowledge, and that it is common
knowledge that knowledge leads to belief.  $R_3$ lets us expand out
common knowledge as unbounded iterated knowledge.  $R_4$ states that
knowledge of a proposition implies that the proposition holds.  $R_5$
to $R_{10}$ provide for a more restricted form of reasoning for
propositions that are common knowledge, unlike propositions that are
known or believed.  $R_{12}$ states that if an agent $s$ communicates
a proposition $\phi$ to $h$, then $h$ believes that $s$ believes
$\phi$.  $R_{14}$ dictates how obligations get translated into
intentions.

\begin{scriptsize}

\begin{mdframed}[linecolor=white, frametitle=Inference Schemata, frametitlebackgroundcolor=gray!25, backgroundcolor=gray!10, roundcorner=8pt]
\begin{equation*}
\begin{aligned}
  &\infer[{[R_{\knows}]}]{\knows(a,t_2,\phi)}{\knows(a,t_1,\Gamma), \ 
    \ \Gamma\vdash\phi, \ \ t_1 \leq t_2} \hspace{10pt} \infer[{[R_{\believes}]}]{\believes(a,t_2,\phi)}{\believes(a,t_1,\Gamma), \ 
    \ \Gamma\vdash\phi, \ \ t_1 \leq t_2} \\
 &\infer[{[R_1]}]{\common(t,\perceives(a,t,\phi) \lif\knows(a,t,\phi))}{}\hspace{6pt}
  \infer[{[R_2]}]{\common(t,\knows(a,t,\phi)
    \lif\believes(a,t,\phi))}{}\\
  &\infer[{[R_3]}]{\knows(a_1, t_1, \ldots
    \knows(a_n,t_n,\phi)\ldots)}{\common(t,\phi) \ t\leq t_1 \ldots t\leq
    t_n}\hspace{10pt}
  \infer[{[R_4]}]{\phi}{\knows(a,t,\phi)}\\
  & \infer[{[R_5]}]{\common(t,\knows(a,t_1,\phi_1\lif\phi_2))
    \lif \knows(a,t_2,\phi_1) \lif \knows(a,t_3,\phi_2)}{}\\
& \infer[{[R_6]}]{\common(t,\believes(a,t_1,\phi_1\lif\phi_2))
    \lif \believes(a,t_2,\phi_1) \lif \believes(a,t_3,\phi_2)}{}\\
& \infer[{[R_7]}]{\common(t,\common(t_1,\phi_1\lif\phi_2))
    \lif \common(t_2,\phi_1) \lif \common(t_3,\phi_2)}{} \\
& \infer[{[R_8]}]{\common(t, \forall x. \  \phi \lif \phi[x\mapsto
  t])}{} \hspace{18pt}
  \infer[{[R_9]}]{\common(t,\phi_1 \liff \phi_2 \lif \neg
    \phi_2 \lif \neg \phi_1)}{}\\
& \infer[{[R_{10}]}] {\common(t,[\phi_1\land\ldots\land\phi_n\lif\phi]
  \lif [\phi_1\lif\ldots\lif\phi_n\lif\psi])}{}\\
&\infer[{[R_{12}]}]{\believes(h,t,\believes(s,t,\phi))}{\says(s,h,t,\phi)}
\hspace{18pt}\infer[{[R_{13}]}]{\perceives(a,t,\happens(\action(a^\ast,\alpha),t))}{\intends(a,t,\happens(\action(a^\ast,\alpha),t'))}\\
&\infer[{[R_{14}]}]{\knows(a,t,\intends(a,t,\chi))}{\begin{aligned}\ \ \ \ \believes(a,t,\phi)
 & \ \ \
 \believes(a,t,\ought(a,t,\phi, \chi)) \ \ \ \ought(a,t,\phi,
 \chi)\end{aligned}}
\end{aligned}
\end{equation*}
\end{mdframed}
\end{scriptsize}

\small
\bibliographystyle{named}
\bibliography{main72,naveen}
\end{document}